\documentclass[conference]{IEEEtran}
\usepackage{times}

\usepackage[numbers]{natbib}
\usepackage{multicol}
\usepackage[bookmarks=true]{hyperref}
\usepackage{graphicx}   
\usepackage{subfigure} 
\usepackage{color}

\begin{document}

\title{What Can I Do Around Here? Deep Functional Scene Understanding for Cognitive Robots}

\author{Chengxi Ye, Yezhou Yang, Cornelia Ferm\"{u}ller, Yiannis Aloimonos
\\
Computer Vision Lab, University of Maryland, \{cxy, yzyang, fer, yiannis\}@umiacs.umd.edu}



%

\maketitle

\begin{abstract}

For robots that have the capability to interact with the physical environment through their end effectors, understanding the surrounding scenes is not merely a task of image classification or object recognition. To perform actual tasks, it is critical for the robot to have a functional understanding of the visual scene. Here, we address the problem of localizing and recognition of functional areas from an arbitrary indoor scene, formulated as a two-stage deep learning based detection pipeline. A new scene functionality testing-bed, which is complied from two publicly available indoor scene datasets, is used for evaluation. Our method is evaluated quantitatively on the new dataset, demonstrating the ability to perform efficient recognition of functional areas from arbitrary indoor scenes. We also demonstrate that our detection model can be generalized onto novel indoor scenes by cross validating it with the images from two different datasets.

\end{abstract}

\IEEEpeerreviewmaketitle

\section{Introduction}

Every day, human beings interact with the surrounding environments
 via perception and action. Through interacting with our environments, we gradually draw on our understanding of the functions that areas of the scene bear. For example, a round knob like object indicates that an action of spherical grasping and perhaps turning could be applied onto, and a faucet like entity with metallic texture indicates that an action of turning on or off water could be applied onto. As human beings, we recognize such functional areas in our daily environments with vision, so we can perform various of actions in order to finish a task. As robots (such as Baxter) begin to collaborate with humans in domestic environments, they will also need to recognize functional areas in order to develop a \textbf{functional understanding} of the scene.

\begin{figure}[!ht]
\centering
\includegraphics[height=1.4in]{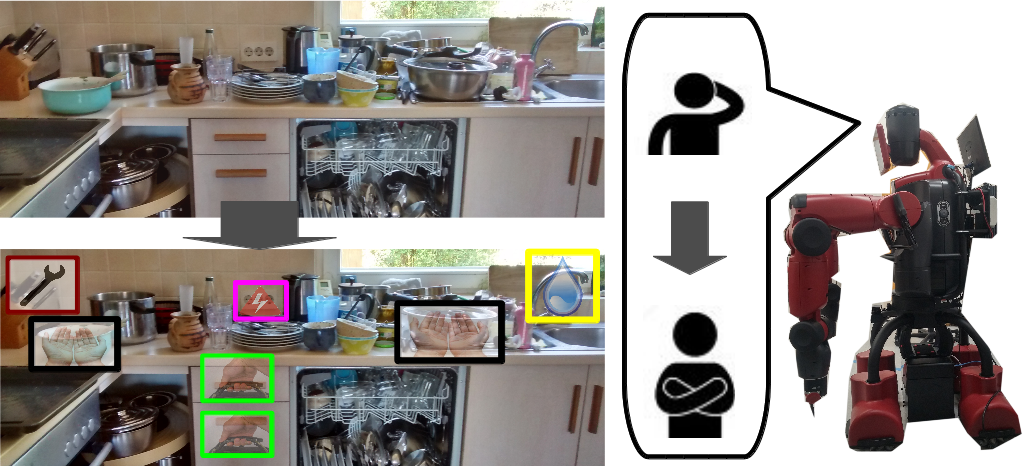}
\caption{The purpose of our system. Through supervised learning, the robot can take appropriate actions to interact with the world by visually observing the environment.}
\label{fg:kit}
\end{figure}

Imagine a Baxter robot with a mobile base enters an arbitrary kitchen, trying to clean up the mess from a dinner party (Fig.~\ref{fg:kit}). Before the robot can perform any dexterous movements, the robot needs to recognize functional areas of the scene such as where it can indicate actions like getting power for a vacuum machine, getting water, etc. Modern Computer Vision technologies allows robots to recognize objects from a known set of hundreds of categories. However, in the previous situation, a brutal applying of object detectors either does not suffice (for example, detecting cabinet does not indicate where and how to open), or is an overkill (for example, it is not necessary to differentiate between paper towel with towel as long as the robot understands a pinch grasp could be applied to get them). Instead, we argue that more importantly the robot needs to know which part of the scene corresponds to which functionality, or to address the inevitable self-questioning: what can I do around here?  As Gibson remarked, ``If you know what can be done with a[n] object, what it can be used for, you can call it whatever you please'' \cite{gibson1977theory}.
      
In this paper, we address the novel problem of {\bf localizing and identifying functional areas of an arbitrary indoor scene}, so that a robot can have a functional understanding of the visual scene in order to perform actions accordingly, and generalize this knowledge into novel scenes. Example outputs of the presented system are shown in Fig.~\ref{fg:kit}: without recognizing the big salad bowl, the robot understands that a two-hand raise-and-move action (black bounding boxes) can be applied in a specific area of the kitchen; without seeing a specific type of handle bar before. However, from its appearance it should be able to infer that a wrap grasp could be applied to pull the bar (green bounding boxes), etc.

Knowing ``what can I do here'' can serve as the first step for many applications in Robotics. For example, the presented pipeline could serve as a functional attention mechanism at the first glance of a novel scene for the robot. The output of the system can guide robot motion planning to move towards target functional area. Dexterous actions can then be planned in these attended areas. Also the usage of our system is not limited in the field of robotics. One potential application could be to provide verbalized functional instructions for the blind. 

From a computer vision and robotic vision perspective, recognizing functional areas from a static scene (in our experiments, we use static images to simulate the scene seen by a robot), is a challenging task since areas with unique visual appearance may indicate the same functionality. Also, unlike concepts such as objects, functionality itself can only be meaningful while a certain action could be applied on or leads to a certain consequence. Therefore we need an ontology for scene functionality. Furthermore, our goal is to provide a general functional scene understanding model for unconstrained indoor scene from real scenarios, and we have to overcome the huge variance naturally imposed from real world image.

The main contributions of this paper are as follows: 1) a scene functional area ontology for indoor domain; 2) the first two stage, deep network based recognition approach is developed and presented for scene functional understanding, which is proved to be effective and efficient on two scene functional area datasets that are augmented from publicly available datasets; 3) the first scene functionality dataset is compiled and made publicly available, which contains more than {\bf 100,000} annotated training samples and two sets of testing images from different indoor scenes\footnote{Available from \url{https://www.umiacs.umd.edu/~yzyang/DFSU}}.
.

\section{Related Work}

We will focus our review on studies of  three  major concepts, which we consider are most closely related to this work: \emph{a)} studies that evolve around the idea of the concept of affordance in Robotics; \emph{b)} studies aiming to infer or reason beyond appearance, such as physics or action intention and \emph{c)} deep network based visual recognition approaches. 

{\bf Affordance related: }
The affordance of objects has been studied in Computer Vision and Robotics communities. Early work sought a function-based approach to object recognition for 3D CAD models of objects, for example chairs \cite{stark1994function}. 
From the computer vision community, \cite{kjellstrom2011visual} classifies human hand actions considering the objects being used, Grabner et al. \cite{grabner2011makes} detects ``sittable'' areas from 3D data. \cite{pieropan2013functional} represents object directly in terms of their interaction with human hands. More recently, \cite{myers2015affordance} attempts to learn the affordance of tool parts from geometric features obtained from RBGD images using different classification approaches. They showed that the concept of affordance has generalization potential.

Affordances might also be considered as a subset of object attributes, which have been shown to be powerful for object recognition tasks as well as transferring knowledge to new categories. 
In the robotics community, the authors of \cite{sun2013attribute} identify color, shape, material, and name attributes of objects selected in a bounding box from RGB-D data. \cite{hermans2013learning} explored, using active manipulation of different objects, the influence of the shape, material and weight in predicting good ``pushable'' locations. \cite{aldoma2012supervised} used a full 3D mesh model to learn so-called 0-ordered affordances that depend on object poses and relative geometry. Koppula et al. \cite{koppula2013learning} view affordance of objects as a function of interactions, and jointly model both object interactions and activities via a Markov Random Field using 3D geometric relations (`on top', `below' etc.) between the tracked human and object as features.

{\bf Reason Beyond Appearance: }
Many recent approaches in Robotics and Computer Vision aim to infer physical properties beyond appearance models from visual inputs. \cite{xie2013inferring} proposes that implicit information, such as functional objects, can be inferred from videos. \cite{yezhou2013cvpr} used stochastic tracking and graph-cut based segmentation to infer manipulation consequences beyond appearance. \cite{zhu2015understanding} takes a task-oriented viewpoint and models objects using a simulation engine.  \cite{pham2015towards} is the first work in Computer Vision that simulates action forces. More recently, \cite{yang2015grasp} proposes that the grasp type, which is recognized using CNNs, reveals the general category of action the human is intending to do next.

{\bf Deep Network based Visual Recognition: }
Recently, studies in neural network have found that convolutional neural networks (CNN) can produce near human level image classification accuracy\cite{krizhevsky2012imagenet}, and related work has been used to various visual recognition tasks such as scene labeling \cite{farabet2013learning} and object recognition\cite{girshick2014rich}. For other closely related technical work that our approach builds upon, we embed the survey along the presentation of our approach in Sec.~\ref{sec:app}

\section{Scene Functionality Ontology}
\label{sec:ontology}

\begin{figure}[!ht]
\centering
\includegraphics[height=2.5in]{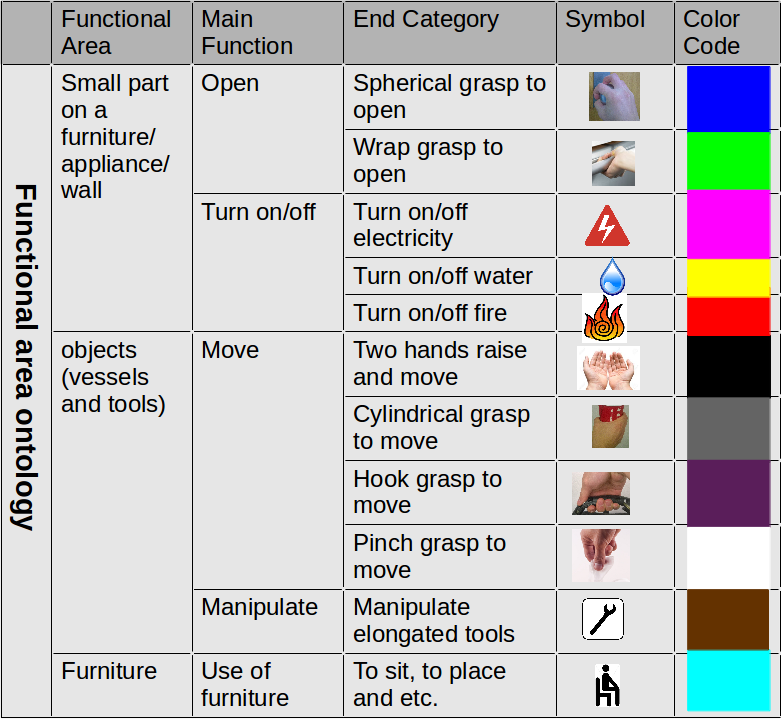}
\caption{Functionality ontology considered in this work.}
\label{fig:ontology}
\end{figure}

A number of functionality ontologies have been proposed in  several areas of research, including robotics, developmental medicine, and biomechanics, each focusing on different aspects of the action. In a recent survey, Worgotter et al. \cite{worgotter2013simple} attempts to categorize manipulation actions into less than 30 fundamental types based on hand-object relations. In this work, we study the common set of actions a robot with end effectors could perform in a domestic indoor environment and use hierarchical categorization of functionality into eleven functional area end categories. 

First we distinguish, according to the most commonly used classification (based on the functional area location), into ``Small part of furniture/appliance/wall'', ``Objects (vessels and tools)'' and ``Furniture''. For the ``Small part of furniture/appliance/wall'' category, we first consider functional areas that could be used to physically open a container, such as a cabinet handle or an oven handle. Based on the main grasping type, we further differentiate the ``to open'' category into two end categories, which are ``spherical grasp (turn) to open'' and ``wrap grasp (drag) to open''.  For the ``to turn on/off'' sub-category, according to intended media, we further differentiate it into three end-categories: ``turn on/off fire'', ``turn on/off water'' and ``turn on/off electricity''. 

Secondly, for the ``Objects (vessels and tools)'' sub-category, we first consider functional areas that are used to move around the appliance. Based on the shape of the functional area and number of hands needed, we further differentiated it into four end categories: ``two-hands raise-and-move'', ``cylindrical grasp to move'', ``hook grasp to move'' and ``pinch grasp to move''. Also, for areas contain elongated tools, we categorize it into end category, which is ``manipulate elongated tools''. 

Lastly, we consider the use of other whole furniture, and categorize it into an end category ``to sit, to place and etc.'', which contains a heavily studied functionality in Computer Vision ``sittable'' \cite{grabner2011makes}.  Fig.~\ref{fig:ontology} illustrates the considered functionality hierarchy and end-categories. We also design different color code for each functional area to facilitate visualization in our detection pipeline.

\section{Our Approach}
\label{sec:app}

In this section, we first briefly summarize the basic concepts involved in an end-to-end visual detection pipeline through Convolutional Neural Networks (CNN), and then we present our implementation of the recognition pipeline which is tailored for the task of scene functional understanding.  

\begin{figure}
\centering
\subfigure[]{\includegraphics[width=1.5in]{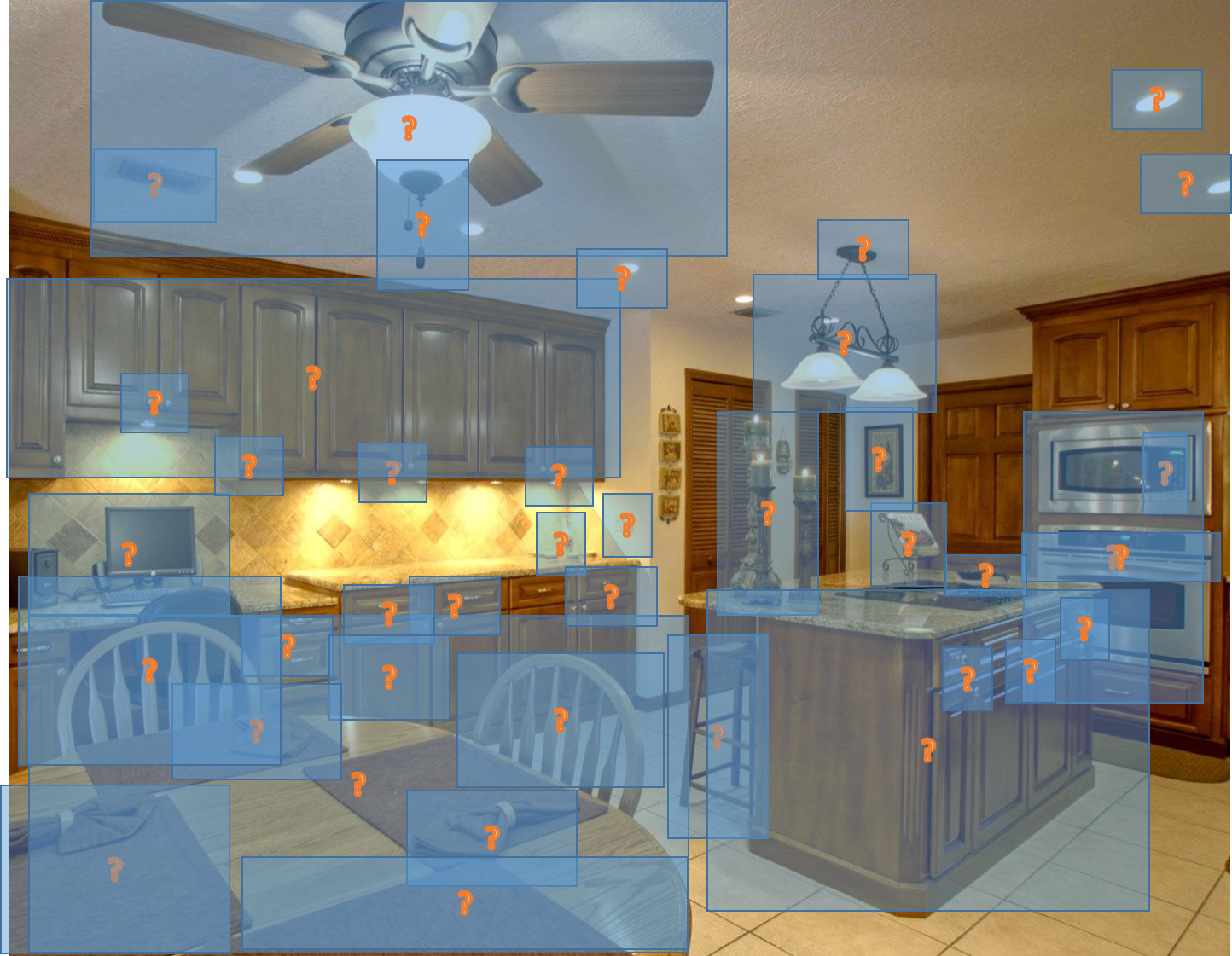}}
\subfigure[]{\includegraphics[width=1.2in]{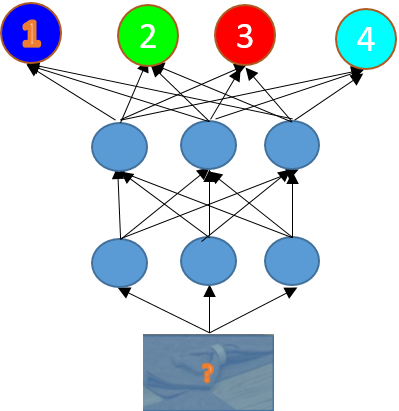}}
\subfigure[]{\includegraphics[width=2.0in]{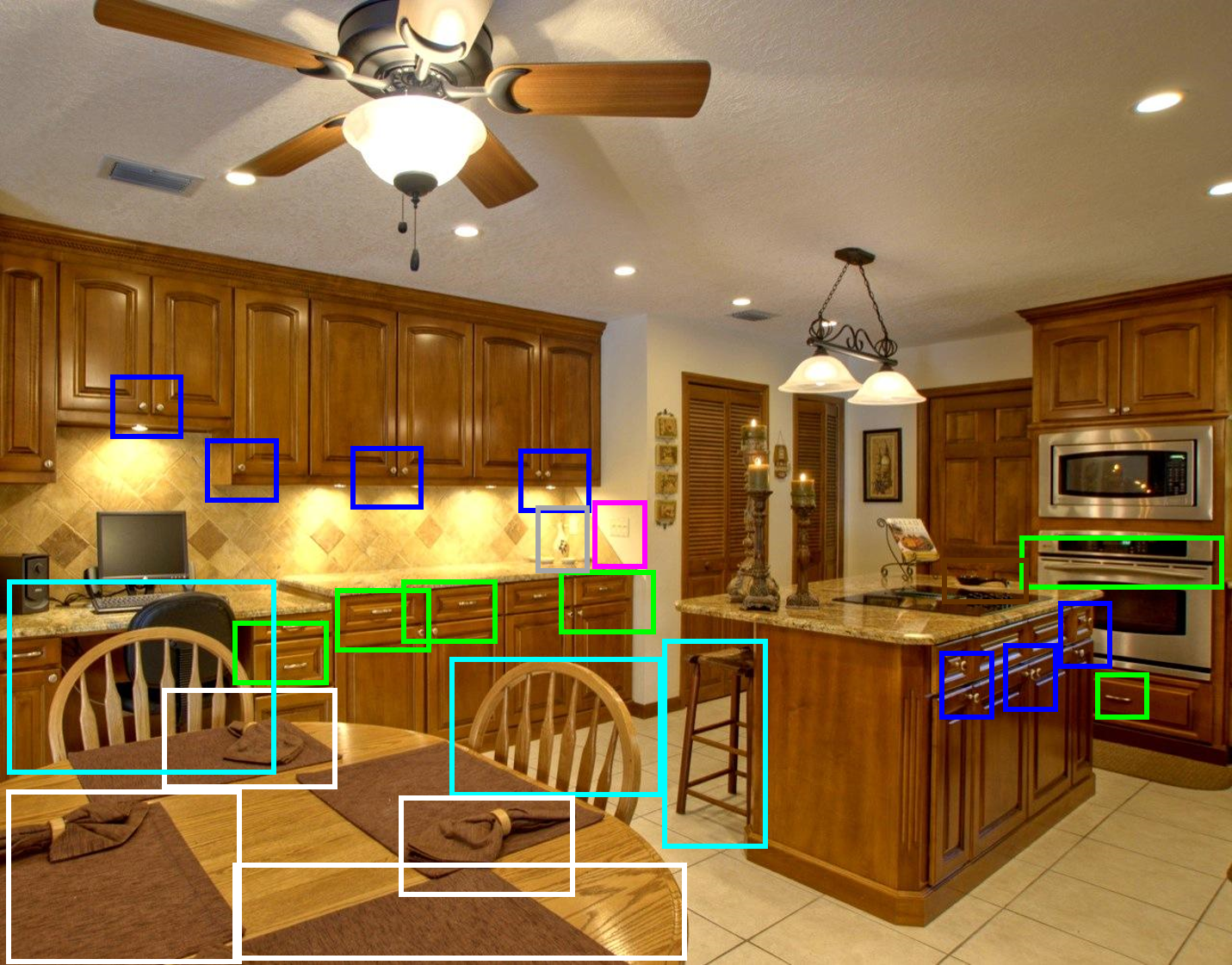}}
\caption{ (a) An area proposal pipeline provides high quality category independent candidate regions. (b) A neural network takes each of the regions and predict the most likely function. (b) System output following color code from Fig.~\ref{fig:ontology}.}
\label{fig:pipeline}
\end{figure}

The input to our system is a static image of an indoor scene (we use kitchen scenes for training), which simulates a scene that a domestic robot assistant, equipped with a normal RGB camera, could observe. 
Fig.~\ref{fig:pipeline} depicts the key steps of our system. First, the system applies an attention mechanism to guide the robot to a set of areas (represented as bounding boxes) that potentially contain functional meaningful areas as depicted in Fig.~\ref{fig:pipeline} (a). Then, a deep neural network is trained to take each of these areas as input and output a belief distribution over the ontology of scene functional areas as depicted in Fig.~\ref{fig:pipeline} (b). In Fig ~\ref{fig:pipeline} (c), the final output of our system is visualized. Bounding boxes are used to indicate the detected location of the functional area and different color is used to indicate the function predicted. For example, the blue boxes around the cabinet's door handles indicate the robot could apply ``spherical grasp to open'' and the green boxes around the oven handle indicates the robot could apply ``wrap grasp to open''. Please refer to Fig.~\ref{fig:ontology} for a full list of functionality considered and their color codes.

\subsection{Functional Region Proposal}

Given an input image, there exists an astronomical amount of possible candidate regions that could contain functional areas. But human visual system has limited processing power. It allocates its resources by limiting to a small amount of attention areas based on low level vision features such as texture, contrast and color. 

Recently, there is a variety of algorithms are presented that use low level vision features to train attention models and output category independent region proposals\cite{endres2010category,alexe2012measuring,uijlings2013selective}. These pipelines combine a diverse amount of visual features, such as color, intensity, texture and edge information, to select areas from static images that are likely to be semantically meaningful. In the first stage of our system, we adopted one of the state-of-the-art visual attention methods called selective search\cite{uijlings2013selective} to propose a set of bounding boxes from an input static scene. 

Selective search over-segments an input image and then hierarchically merges similar neighboring regions to reduce the amount of candidate regions. The similarity measure is based on a variety of criteria, such as size, color, texture. Different from exhaustive search, selective search aim to provide higher quality but a small number of candidate regions. These regions serve as candidates to the next stage processing. Though selective search has been proven to be effective, it may still output background areas. Thus in the following steps, together with the eleven functional categories introduced in Sec.~\ref{sec:ontology}, we also include a ``background'' category. 

\subsection{CNN for feature extraction}

Biologically inspired learning algorithms such as the artificial neural network were heavily used in the 1990s \cite{lecun1998gradient}. Artificial neural networks computationally simulate the activities of human neurons. Each node in a network acts as an artificial neuron that takes as inputs the outputs of other nodes, and applies a non-linear mapping of the combined inputs as its own output. 
The human cerebral cortex system was estimated to have $\sim $100 billion neuron cells organized in 6
layers\cite{herculano2009human,lui2011development} to conduct different perceptual activities. The complex and deep layered structure may be a key factor to the human perceptual capability.
In the early days, due to limited computational resources and the inherent difficulty in training a layered non-linear system, artificial neural networks used to have much fewer elements ($\sim$$10^{3}$) and layers (2-3). With recent advancements in hardware development and progress in training large scale non-linear functions, developments of artificial neural network witness much improved number of elements($\sim10^{7}$) and depths($>5$). As a result, the recognition power of the corresponding networks have been improved to get closer to human performance\cite{Krizhevsky2012}. 

Convolutional neural network (CNN), which is a specific type of neural network, was developed for computers to simulate the human vision system. Each convolutional element in the network spatial-invariantly combines neighboring feature maps as $y =\sum_{1 \leq i \leq N}{X_{i} \ast w_{i}}+b$, where $X_{i}$ is the input feature map from the $i-th$ neuron from the previous layer. $w_{i}$ is a convolution kernel corresponding to that $i-th$ node that simulates the human receptive field. $\ast$ is a convolution operator and $b$ is a bias to the linear output. The convoluted results using different kernels are then linearly combined as inputs to a non-linear mapping as: 

\begin{equation}
z=f(y)=f(\sum_{1 \leq i \leq Inputs}{X_{i} \ast w_{i}}+b).
\label{eq1}
\end{equation}

The neural network may have other elements such as max/average pooling (locally calculate the max/average elements) to simulate human response to the non-linearly mapped features. The non-linear mapping function $f$ is usually a sigmoid-like function or the rectified linear unit (ReLU) function\cite{Krizhevsky2012}. In this work, we adopt the ReLU function.

To predict the most likely action of an input region, we pass the image patch through a trained neural network. The output features are used as inputs to a softmax function to model the probability $p_j$ of a specific functionality $j$. 

\begin{equation}
p_j=\frac{e^{x_j}}{\sum_{1 \leq i \leq K} {e^{x_i}}}
\label{eq2}
\end{equation}

In this work we adopt two convolutional neural network architectures\cite{chatfield2014return}(VGG-F and VGG-S, differing in the total number of parameters and filter size), which both are pre-trained on the ImageNet dataset~\cite{deng2009imagenet}. Each of them has five convolution layers and three fully connected layers.

\subsection{Network Training}

Training a large network with only a small amount of data could be problematic. Training with insufficient data may lead to a network that overfits the data which has limited generalization power. In the early developments of the human brains, humans gain cognitive capability by being exposed to a vast amount of varied information. The connections between the neurons are pruned during the learning process to adapt to the data and improve efficiency ~\cite{paolicelli2011synaptic}.

One way to adapt a large neural network (pre-trained on millions of samples) to a relatively smaller dataset (with tens of thousands of  training samples) is to fix the network structure and only update the weights of the final mapping to the softmax functions. An even more biologically plausible  strategy\cite{girshick2014rich,chatfield2014return}, that significantly improves the performance, is to simultaneously fine-tune all the parameters in the original neural network.  

In our system, we adopt the fine-tuning strategy and connect the 4096-dimensional output feature vector from pre-trained networks to a (11+1) class (11 functional category introduced in Sec.\ref{sec:ontology} and one ``background'' category) fully connected prediction layer.

There have been many studies on how to train a deep neural network. One simplest approach is to the stochastic gradient descent (SDG) method. For each iteration the algorithm takes a small batch of samples, and evaluate an averaged gradient based on these samples. Parameters are locally optimized in the negative gradient direction to reduce the training loss. A momentum term is introduced to combine the previous estimated gradient to smooth the current gradient estimate. Approaches fall into this paradigm are termed the first order gradient descent approach. This gradient descent approach is a practical and efficient way to reduce the loss when the step size is small enough and when the underlying gradient is non-zero. Remarkably, this approach works well even for non-convex optimization problems such as training the neural network. In practice the training process will converge at a local minimum when proper parameters are chosen. 

\begin{figure}
\centering
\includegraphics[width=3.5in]{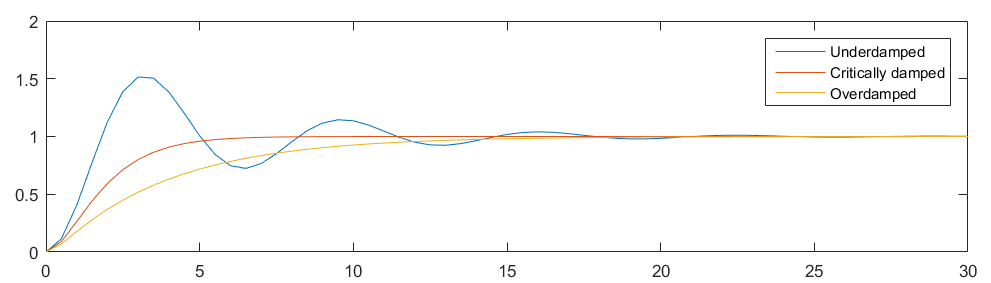}
\caption{ Convergence rates of a 1-d damped spring mass system. Different learning rates lead to different convergence rates. High learning rate causes underdamping, in which case oscillations and slow convergence occur, while low learning rate causes overdamping and slow convergence.}
\label{Damping}
\end{figure}

It can be shown the convergence process with the local convex approximation is equivalent to a high dimensional spring-mass system in a viscous medium \cite{qian1999momentum}. The momentum term amounts to the mass size and accelerates the convergence when set in a proper range. Larger step size amounts to less viscous medium and less damping, in which case the system can be under-damped and will oscillate (shown in  Fig ~\ref{Damping}). In the final stage of training when the parameters are already located near a local minimum, it is practically effective to reduce the step size or reduce the so-called learning rate, in order to magnify the damping amplitude  and accelerates the speed of convergence along directions with larger curvatures.

\section{Experiments}

\subsection{Dataset and Protocols}

Since there does not exist a publicly available dataset for scene functionality analysis, it is necessary to compile a new one to serve as a testing-bed for our pre-mentioned system. 
Without the loss of generality, we set our main testing scenario in a kitchen domain, and we first adopted the subset of kitchen images from a well-known publicly available database (the SUN dataset\cite{xiao2010sun}).

Following the functionality ontology given from Sec.~\ref{sec:ontology}, we manually labeled over 10,000 functional regions from over 500 kitchen photos, each with a category from the eleven entries. We further split the dataset into 90\% for training and 10\% for testing. Additionally, for each training image, we randomly generated $100$ bounding boxes of varied sizes, and treated the patches as ``background'' samples if the Intersection-Over-Union ($IOU=\frac{A \cap B}{A \cup B}$) values of the patch with regard to each labeled functional area are smaller than $0.5$. Following this procedure, we generated around 90,000 ``background'' samples. Overall, we compiled a new dataset which contains more than \textbf{100,000} training samples to serve as initial training data. During the training process, we reserved 2\% of the data for model validation.

The first testing set includes the reserved 10\% of the kitchen images from SUN dataset. To see how well the trained model could generalize onto another dataset, we also randomly selected additional 100 images from anther publicly available dataset (the NYU indoor scene dataset \cite{silberman2011indoor}).  Note that in NYU testing images, there are not only kitchen scenes, but also other indoor scenes, such as office and bedroom. 

\subsection{Training and Testing Details}

We applied the first order gradient descent approach to train the network. The learning rate was initially set to be 0.005 for the first 7 layers of the network and 0.05 for the final layer, which generates input to the 
softmax function. As mentioned before, we reduced the learning rate by a magnitude of 10 after 40 epochs. The significant drop in training loss may be well explained by the fact that smaller learning rate corresponds to higher damping rate, and results in faster convergence in certain high curvature directions. After another 20 epochs, we further reduced the learning rate by another magnitude of 10. We then early stopped the training at 70 epochs. All parameters were chosen from previous literature and empirical testing. 

In the region proposal stage of the testing phase, we used the selective search ``quality'' option and set minimum segment size to be the 1/100*max(height,width). With this minimum segment size, selective search algorithm generates thousands of candidate regions for the classification stage of the pipeline.

\begin{figure}
\centering
\subfigure[]{\includegraphics[width=1.5in]
{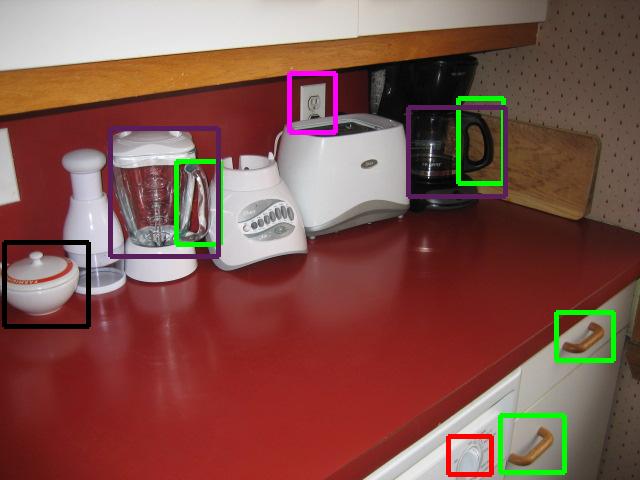}}
\subfigure[]{\includegraphics[width=1.5in]{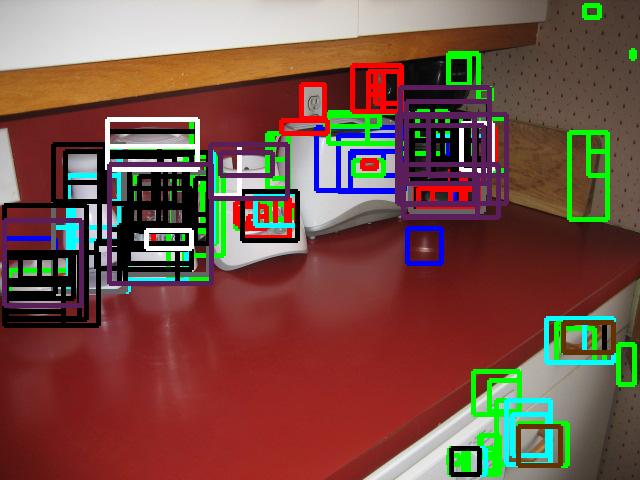}}
\subfigure[]{\includegraphics[width=1.5in]{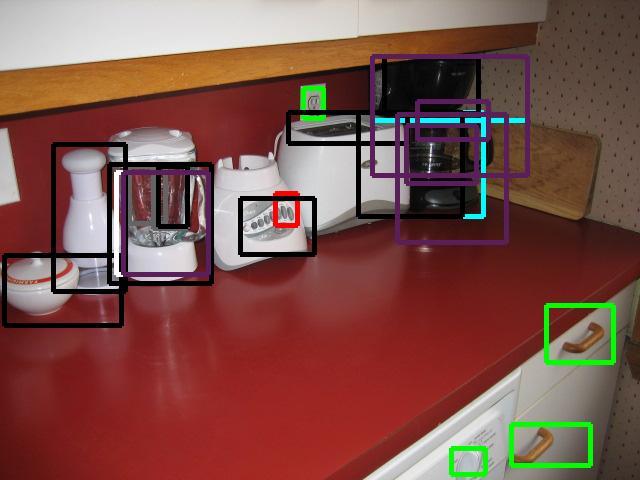}}
\subfigure[]
{\includegraphics[width=1.5in]
{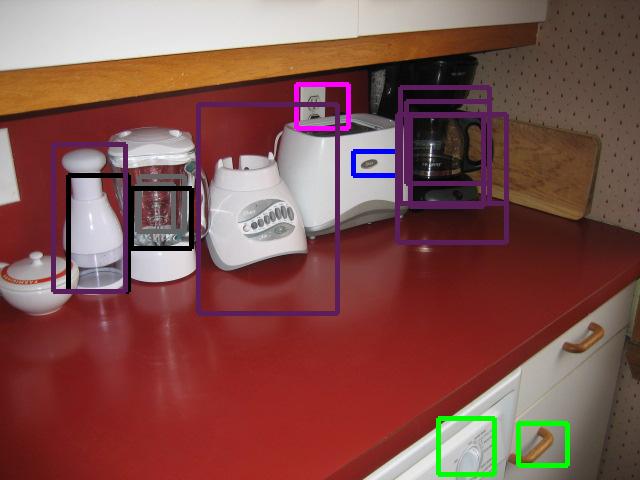}}

\caption{ (a) Ground-truth labeling of a kitchen scene. Detection result with VGG-S network after (b) one round training; (c) two rounds of training; (d) three rounds of training.}
\label{S Networks}
\end{figure}

It is also worth to mention that the system's output after one round of training has large number of false positives (shown in Fig.~\ref{S Networks} (b)) and the precision rate is around 5\%. To improve the performance, we adopted the ``hard negative mining'' technique which is widely used for training an SVM \cite{felzenszwalb2010object}. Specifically, the initial trained network is first applied onto the training images, and the falsely detected regions (false positives and false negatives) were collected as ``hard'' examples. These ``hard'' samples were included as training data into the next round of training. At the end, around $250,000$ images were used in the third round of training. This strategy significantly reduces the false positive rate, and yields a model with the best performance (Fig.~\ref{ROC}).

\subsection{Evaluation metrics}

To compare the performances from different training procedures, we adopted the following evaluation metrics: precision, recall rate and the F-1 score. When calculating these metrics, we treated a detected bounding box as true positive if 1) it has a correct prediction and 2) its IOU value with a labeled box is bigger than $0.5$. The metrics were further calculated following their definitions: $precision=\frac{\#true positive}{\#false positive + \#true positive}$, $recall=\frac{\#true positive}{\#true positive + \#false negative}$ and $F1=\frac{precision \cdot recall}{precision + recall}$. 

\subsection{Experimental results}

\begin{table*}[tbp]
\caption{Precision and Recall for Different Networks}
\centering
\label{PrecRec}
\begin{tabular}{|r|r|r|r|r|r|r|}
\hline
                                                          & \begin{tabular}[c]{@{}r@{}}Precision (SUN)\end{tabular} & \begin{tabular}[c]{@{}r@{}}Recall (SUN)\end{tabular} & \begin{tabular}[c]{@{}r@{}}F1 score (SUN)\end{tabular} & \begin{tabular}[c]{@{}r@{}}Precision  (NYU)\end{tabular} & \begin{tabular}[c]{@{}r@{}}Recall  (NYU)\end{tabular} & \begin{tabular}[c]{@{}r@{}}F1 score  (NYU)\end{tabular} \\ \hline
\begin{tabular}[c]{@{}r@{}}VGG-S One round\end{tabular}  & 5.26\%                                                     & 23.92\%                                                & 0.0862                                                    & 1.69\%                                                     & 50.25\%                                                 & 0.0327                                                    \\ \hline
\begin{tabular}[c]{@{}r@{}}VGG-S Two rounds\end{tabular} & 16.18\%                                                    & 15.05\%                                                & 0.1559                                                    & 8.48\%                                                     & 15.58\%                                                 & 0.1098                                                    \\ \hline
\begin{tabular}[c]{@{}r@{}}VGG-S Three rounds\end{tabular} & 31.52\%                                                    & 11.58\%                                                & 0.1694                                                    & 23.82\%                                                    & 15.53\%                                                 & 0.188                                                     \\ \hline
\begin{tabular}[c]{@{}r@{}}VGG-F One round\end{tabular}  & 4.78\%                                                     & 25.08\%                                                & 0.0803                                                    & 1.29\%                                                     & 49.75\%                                                 & 0.0251                                                    \\ \hline
\begin{tabular}[c]{@{}r@{}}VGG-F Two rounds\end{tabular} & 19.89\%                                                    & 9.08\%                                                 & 0.1247                                                    & 13.34\%                                                    & 15.59\%                                                 & 0.1438                                                    \\ \hline
\begin{tabular}[c]{@{}r@{}}VGG-F Three rounds\end{tabular} & 29.74\%                                                    & 11.54\%                                                & 0.1663                                                    & 13.41\%                                                    & 10.05\%                                                 & 0.1149                                                    \\ \hline
\end{tabular}
\end{table*}

\begin{figure}
\centering
{\includegraphics[width=2.2in]{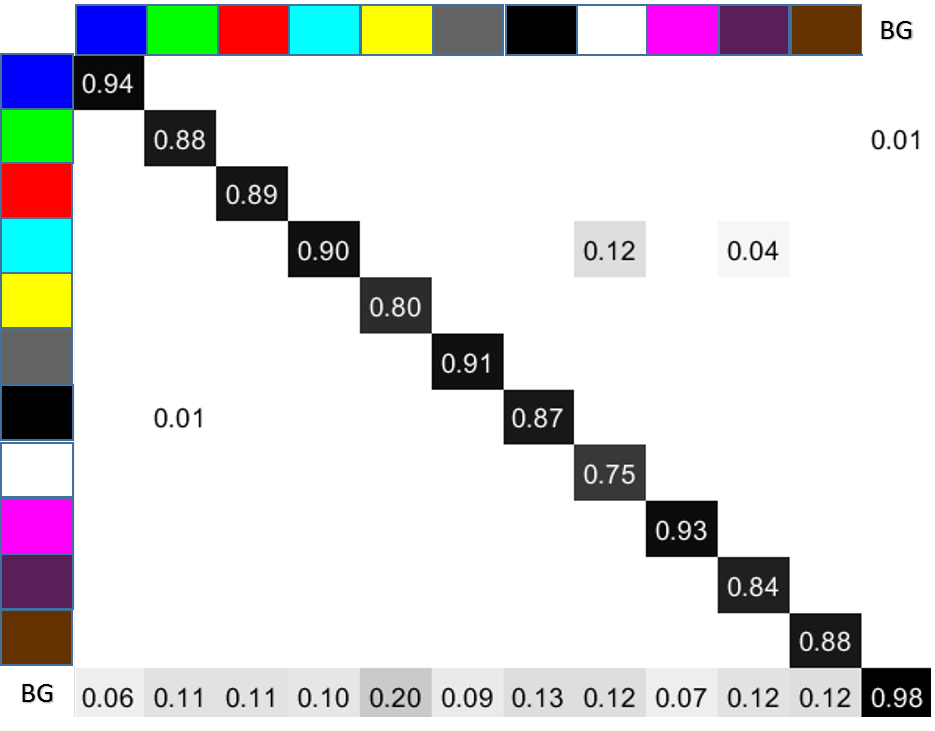}}
\caption{Confusion matrix of the VGG-S network after third round training.}
\label{confmat}
\end{figure}

\begin{figure}
\centering
\subfigure[]
{\includegraphics[height=2.3in]{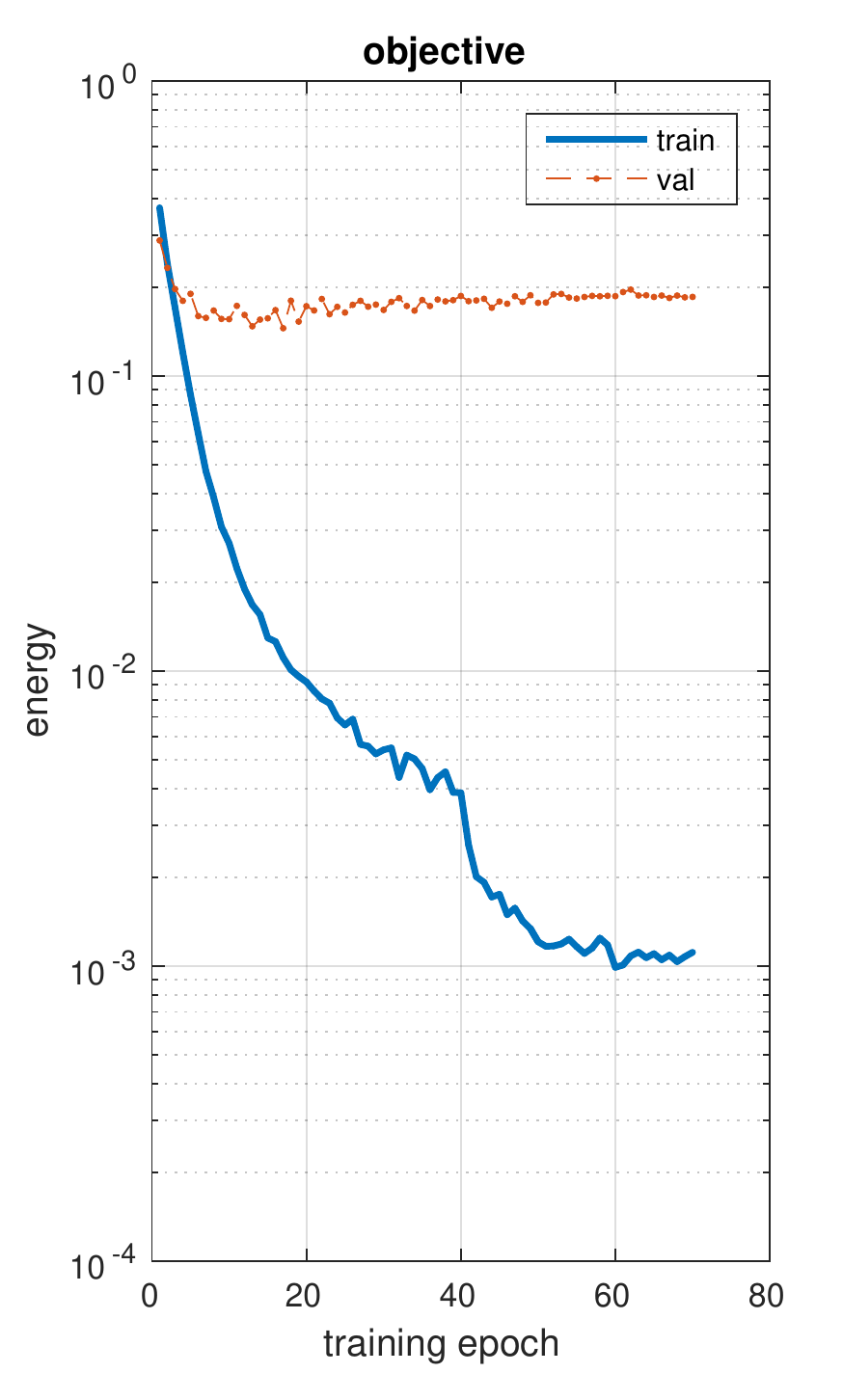}}
\subfigure[]{\includegraphics[height=2.3in]{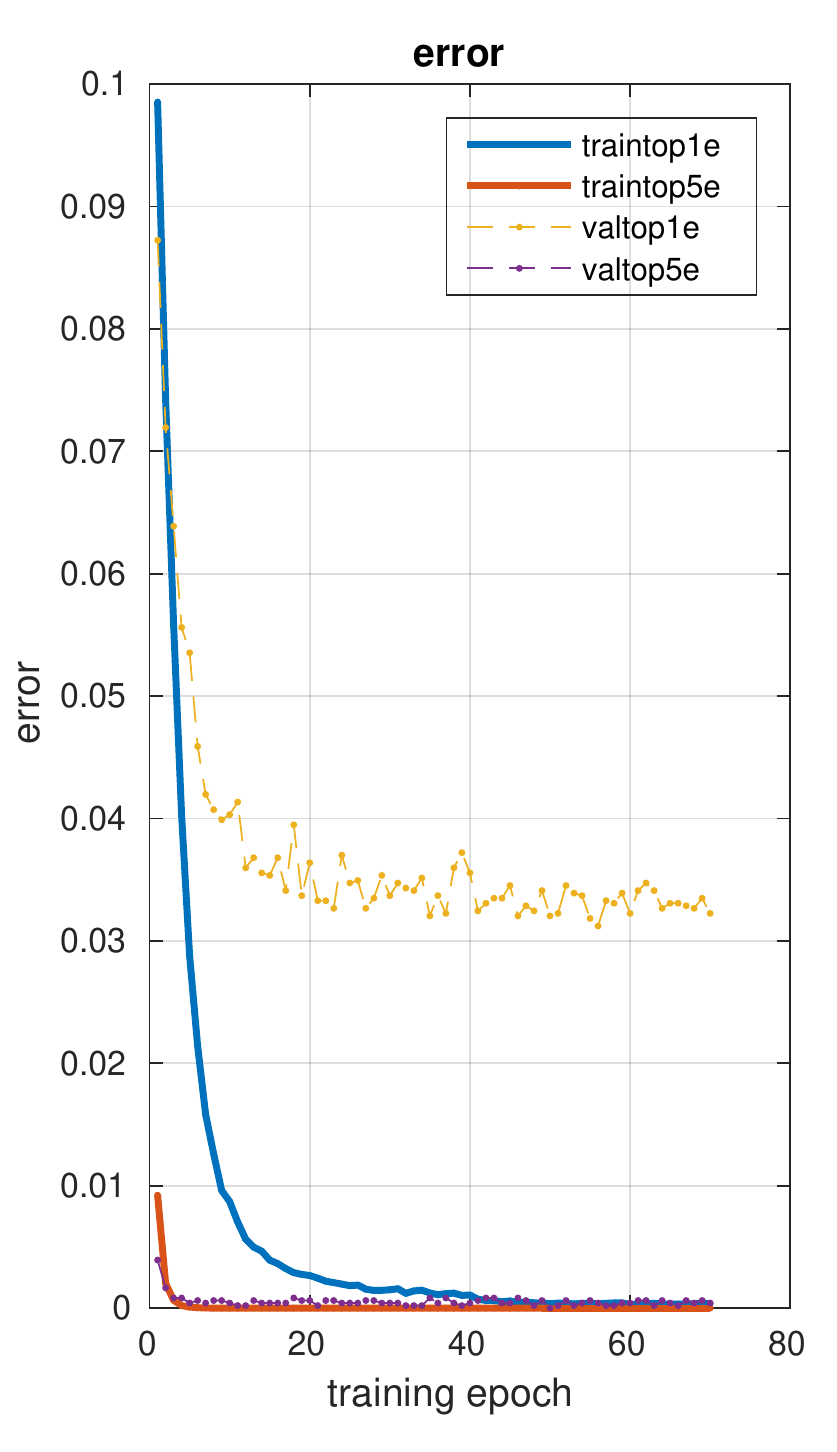}}
\caption{ The loss (a) and the prediction accuracy on validation set (b) of the VGG-S neural network during the third round of training.}
\label{TrainingCurves}
\end{figure}

\begin{figure}
\centering
\subfigure[]
{\includegraphics[width=2.32in]{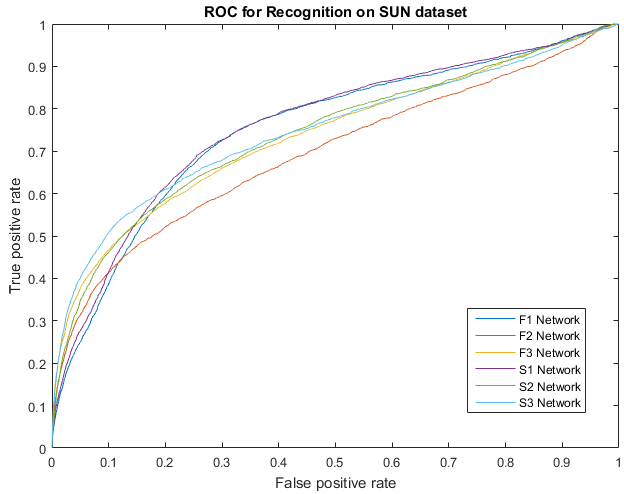}}
\subfigure[]{\includegraphics[width=2.32in]{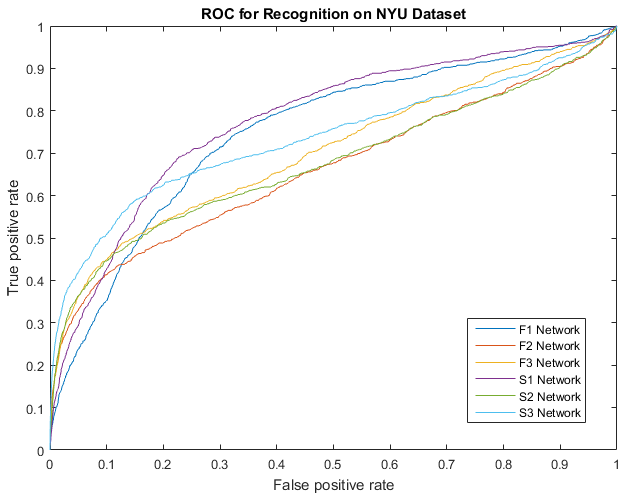}}
\caption{(a) ROC curves of different models on the SUN kitchen dataset. (b) ROC curves of the same set of trained models on NYU indoor scene dataset.}
\label{ROC}
\end{figure}

The ROC curves of different models trained with different rounds can be found in Fig ~\ref{ROC}. In Table~\ref{PrecRec} we report the precision, recall rates and the F1 score of each model on the two testing sets with one to three rounds of training. 
From  Fig. ~\ref{ROC} and Table~\ref{PrecRec}, we can see that,  comparing with the slightly faster but slim VGG-F network, the VGG-S network, whose network architecture is  more complex, achieves better performance at the same recall rate. We also plot the loss curves, top 1-class/5-class error rates curves in the third round training in Fig.~\ref{TrainingCurves}. The corresponding confusion matrix on the validation dataset is plotted in Fig~\ref{confmat}. 

From the Table~\ref{PrecRec}, we can also see that the inclusion of ``hard'' examples from previous rounds of training can significantly improve the detection performance. As a matter of fact, we observe a boost of precision from 5\% to 30\% after three rounds of training. Even though the recall rates may drop after several rounds of training, the F1 score, which takes a trade-off between precision and recall, shows that the VGG-S network with three rounds of training performs best while compared to the VGG-F network. The ROC curves from Fig. ~\ref{ROC} also support this claim. 
Thus, in the rest of experiments, we mainly adopted the model trained using VGG-S architecture after three rounds of training.

\begin{figure*}
\centering
\subfigure[]{\includegraphics[height=1.20in]{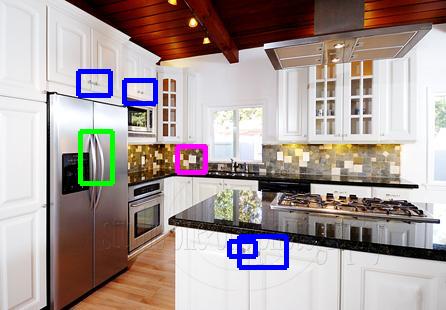}}
\subfigure[]{\includegraphics[height=1.20in]{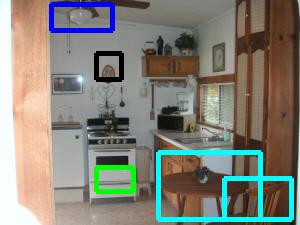}}
\subfigure[]{\includegraphics[height=1.20in]{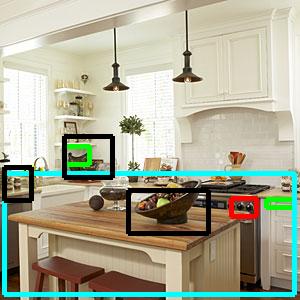}}
\subfigure[]{\includegraphics[height=1.20in]{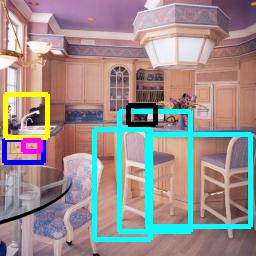}}
\subfigure[]{\includegraphics[height=1.20in]{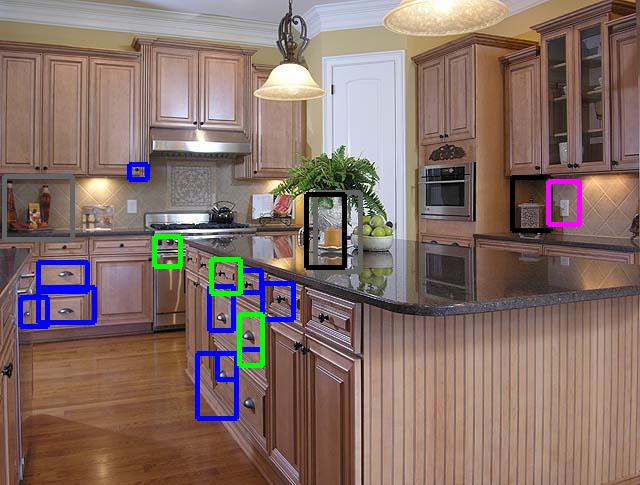}}
\subfigure[]{\includegraphics[height=1.20in]{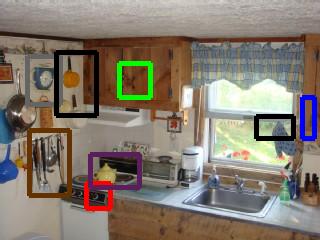}}
\subfigure[]{\includegraphics[height=1.20in]{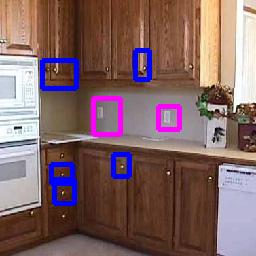}}
\subfigure[]{\includegraphics[height=1.20in]{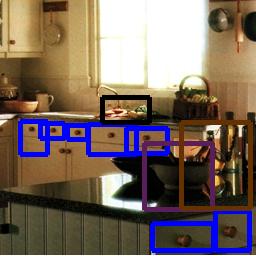}}

\caption{ Sample detection results on the test images from the SUN kitchen dataset, using our system with the VGG-S network after three rounds of training.}
\label{VGG-S 3}
\end{figure*}

Here we want to mention a caveat. Though we had a high labeling standard for compiling the new dataset, there inevitably still exists functional areas that the human labeler failed to label out, such as the ones shown in  Fig ~\ref{S Networks} (a). However, from Fig ~\ref{S Networks} (c,d), we can see that the system is still able to detect these areas as meaningful functional parts. For example, the blender (second left) on the table was not labeled as ``two hands to raise'' in the ground truth, but was successfully detected after three rounds of training on VGG-S network. Using pre-mentioned evaluation metrics, this example was treated as a ``false-positive'', which hurts the model's precision score. 

Furthermore, real world functional areas can be of different shapes and sizes. The labeled area usually deviate from the system proposed area. This phenomenon affected the IOU values, biased them towards lower value and consequently compromised the performance metrics of the system. This indicates that the actual performance of our system is potentially beyond the evaluation metrics reported. 

To better demonstrate the system's performance, we use bounding boxes with different colors to indicate the functional areas detected. The correspondence between the functionality category and the color code can be found in Fig ~\ref{fig:ontology}. 

As mentioned before, we observed that the system trained using VGG-S network after three rounds of training performs best. We used this trained model to demonstrate detection results on the testing images in the SUN kitchen dataset for analysis. From Fig.~\ref{VGG-S 3}, we can see that our system is able to detect meaningful functional areas and output high quality bounding box detections. It can correctly predict intended functionality even when the areas are tiny and ambiguous. It is also interesting to note that the system is able to detect meaningful functionality at some challenging areas of the scene. For example, a bulb (Fig. ~\ref{VGG-S 3})(b) is recognized as a ``spherical grasp to open'' functional area even though the bulb is not labeled with any specific functionality in the training data. Moreover, a round table in the same scene is recognized as ``sittable'', which is physically correct but a rarely used functionality of it. The performance of the system from another aspect supports our intuition that the concept of functional visual areas share similar visual appearance among the same category and is one of the prerequisites for the generalization capability.

Since the functionality ontology is innately general to other indoor domains, we could expect the recognition capability of the trained model to be generalized to different indoor scenes. To validate that the trained model is actually capable of generalizing, we randomly selected 100 scenes from another publicly available NYU indoor dataset and also labeled them with ground-truth functional areas to serve as the second testing set. The new testing set includes static scenes such as office, living room, bedroom and bathroom. We applied the pre-trained detection model from SUN kitchen dataset directly onto this new set of scenes. 

The ROC curves and evaluation scores on the new NYU testing set can be found in Fig~\ref{ROC} (b) and parts of Table ~\ref{PrecRec}. After applying our detection pipeline using the pre-trained model, we achieved comparable performance on the new testing data. 
The metric reported are competitive with the ones in the kitchen scenario. More significantly, the VGG-S model after three rounds of training performs even better on the NYU testing set than on the SUN testing set. 

\begin{figure*}
\centering
\subfigure[]{\includegraphics[width=1.5in]{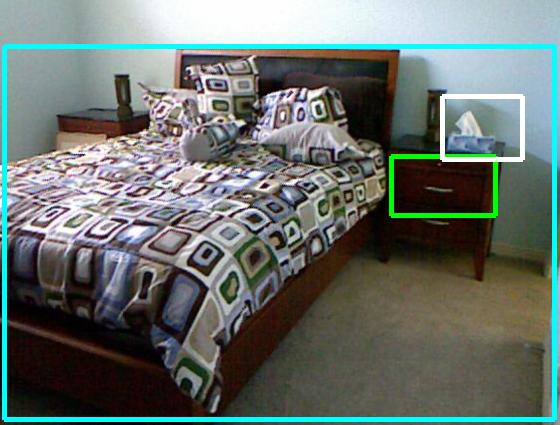}}
\subfigure[]{\includegraphics[width=1.5in]{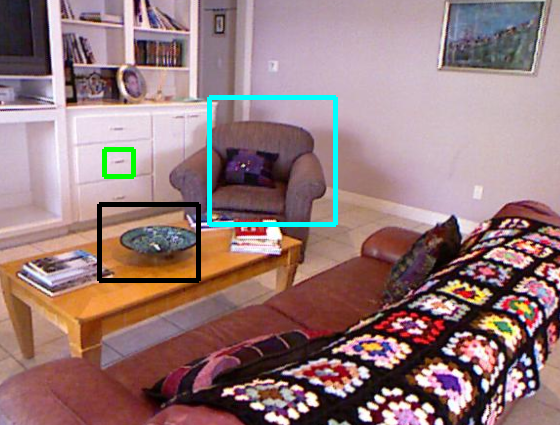}}
\subfigure[]{\includegraphics[width=1.5in]{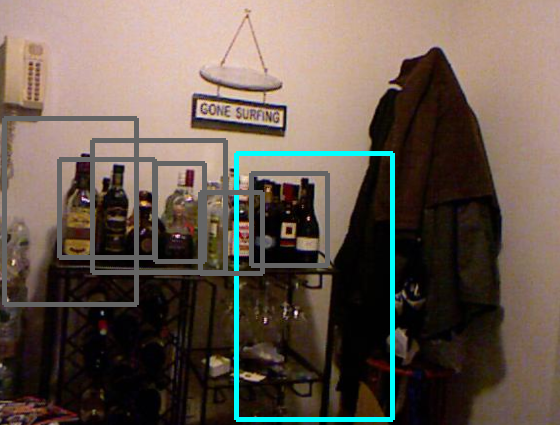}}
\subfigure[]{\includegraphics[width=1.5in]{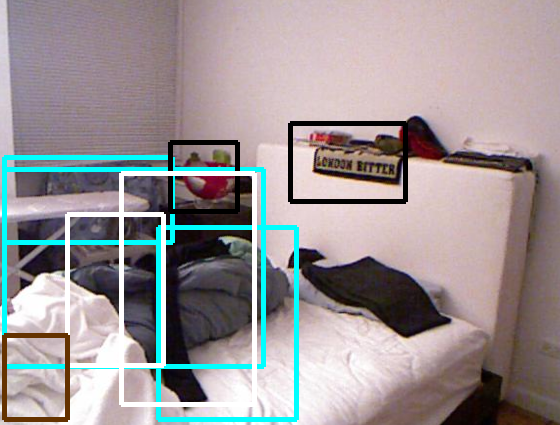}}
\subfigure[]{\includegraphics[width=1.5in]{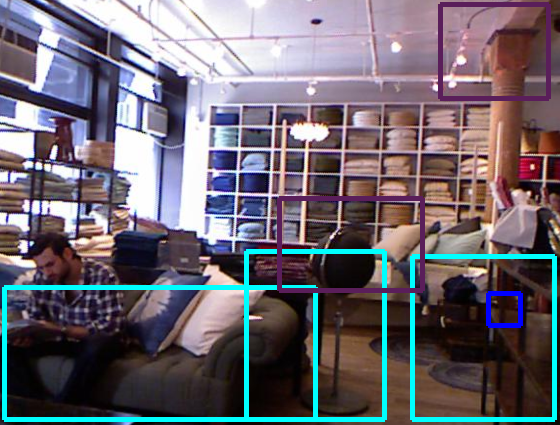}}
\subfigure[]{\includegraphics[width=1.5in]{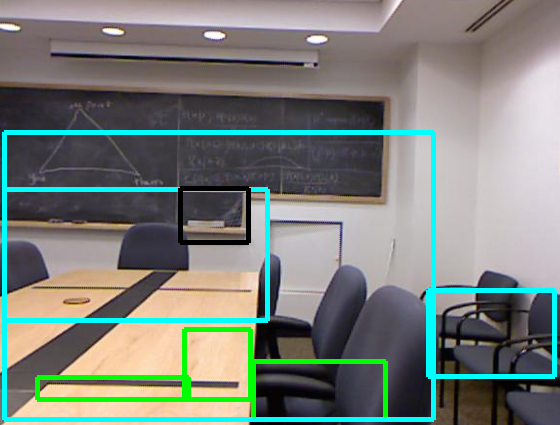}}
\subfigure[]{\includegraphics[width=1.5in]{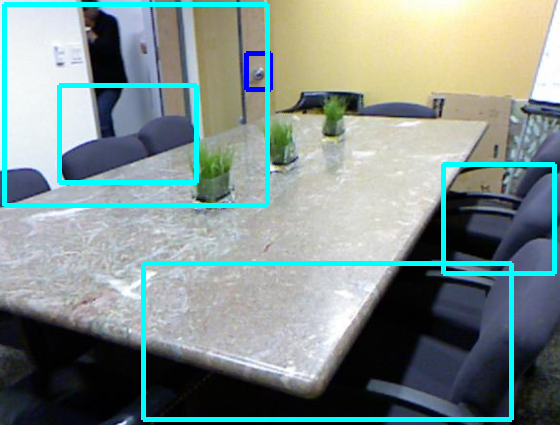}}
\subfigure[]{\includegraphics[width=1.5in]{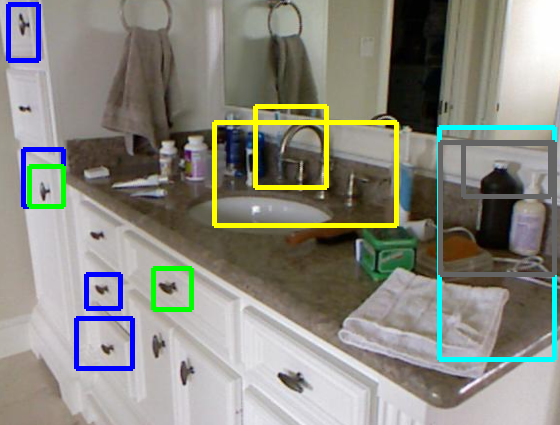}}

\caption{ Sample detection results from the test images of NYU indoor dataset, using our system with the VGG-S network after three rounds of training.}
\label{VGG-S 3 NYU}
\end{figure*}

We also demonstrate the detection results on the NYU testing data using bounding boxes with the same color code (shown in Fig.~\ref{VGG-S 3 NYU}). We can see that, for example, though some objects such as sofa and bed never exist in the kitchen scene, the meaningful functionality ``sittable'' can still be correctly detected (Fig.~\ref{VGG-S 3 NYU}(a,b,d,e)). 
These experimental results on the NYU testing data support our hypothesis that the functional recognition model learned in the kitchen environment is well generalizable to other indoor scenarios. 

\begin{figure*}[!ht]
\centering
\subfigure[]{\includegraphics[width=2.1in]{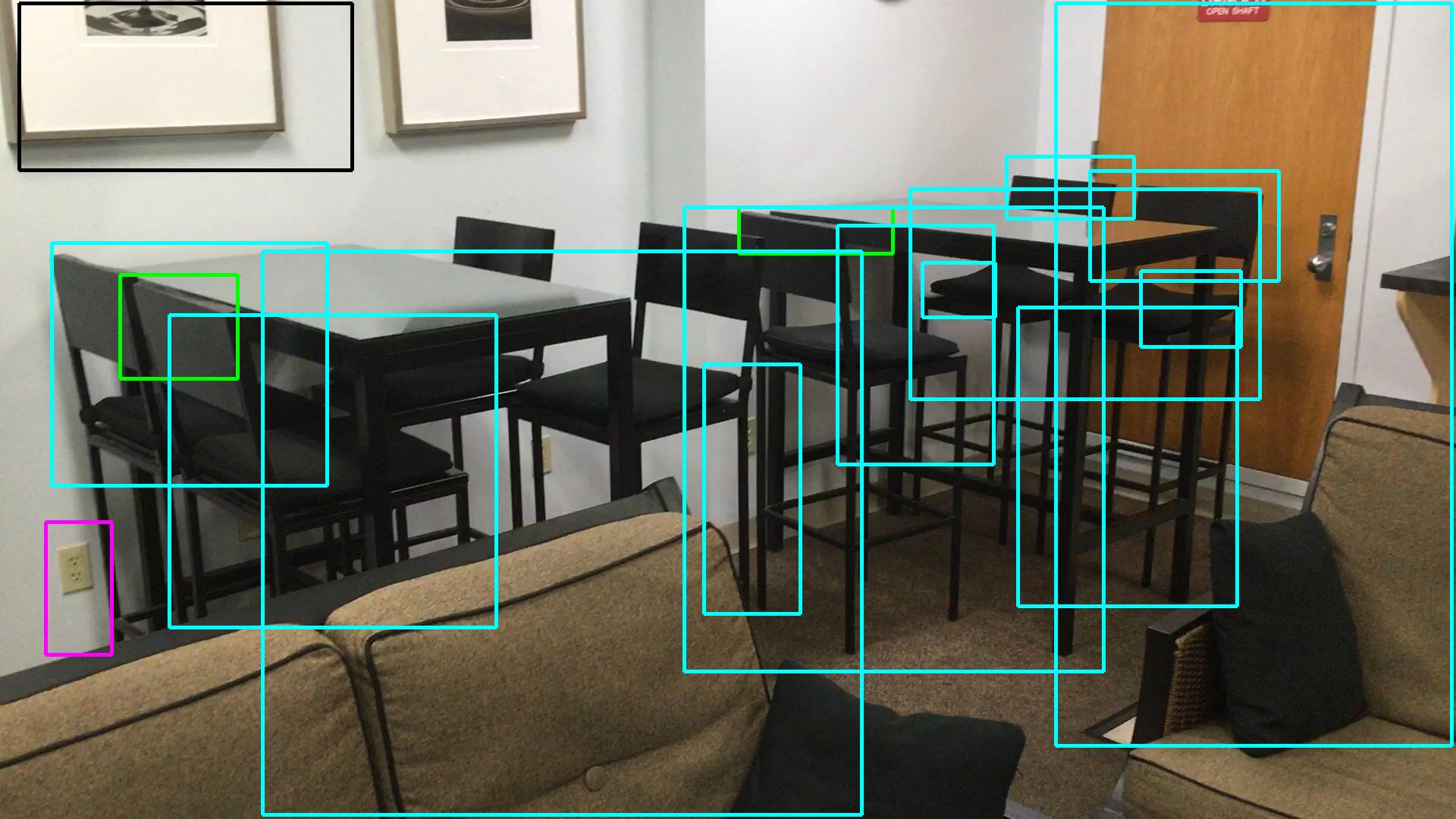}}
\subfigure[]{\includegraphics[width=2.1in]{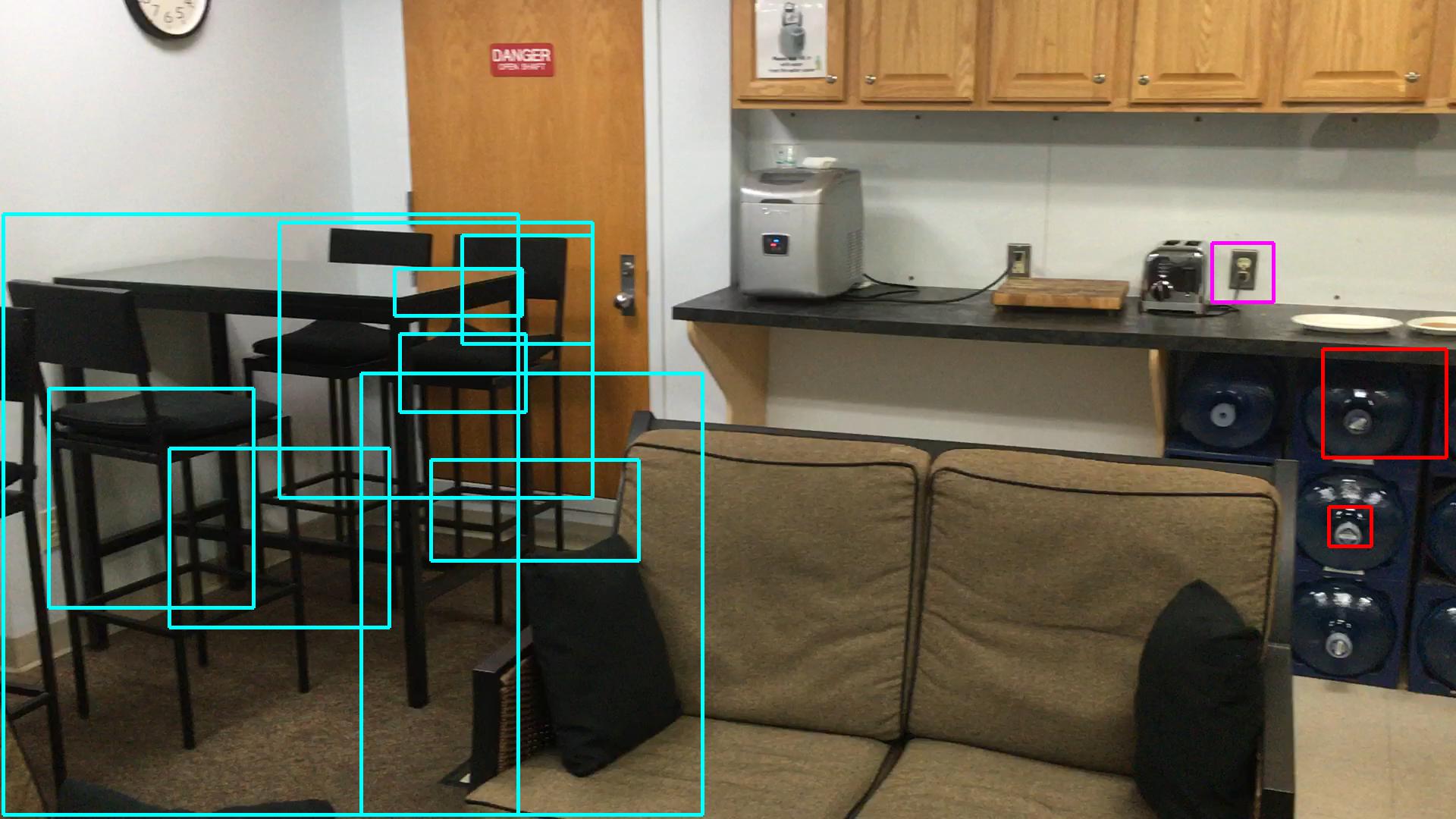}}
\subfigure[]{\includegraphics[width=2.1in]{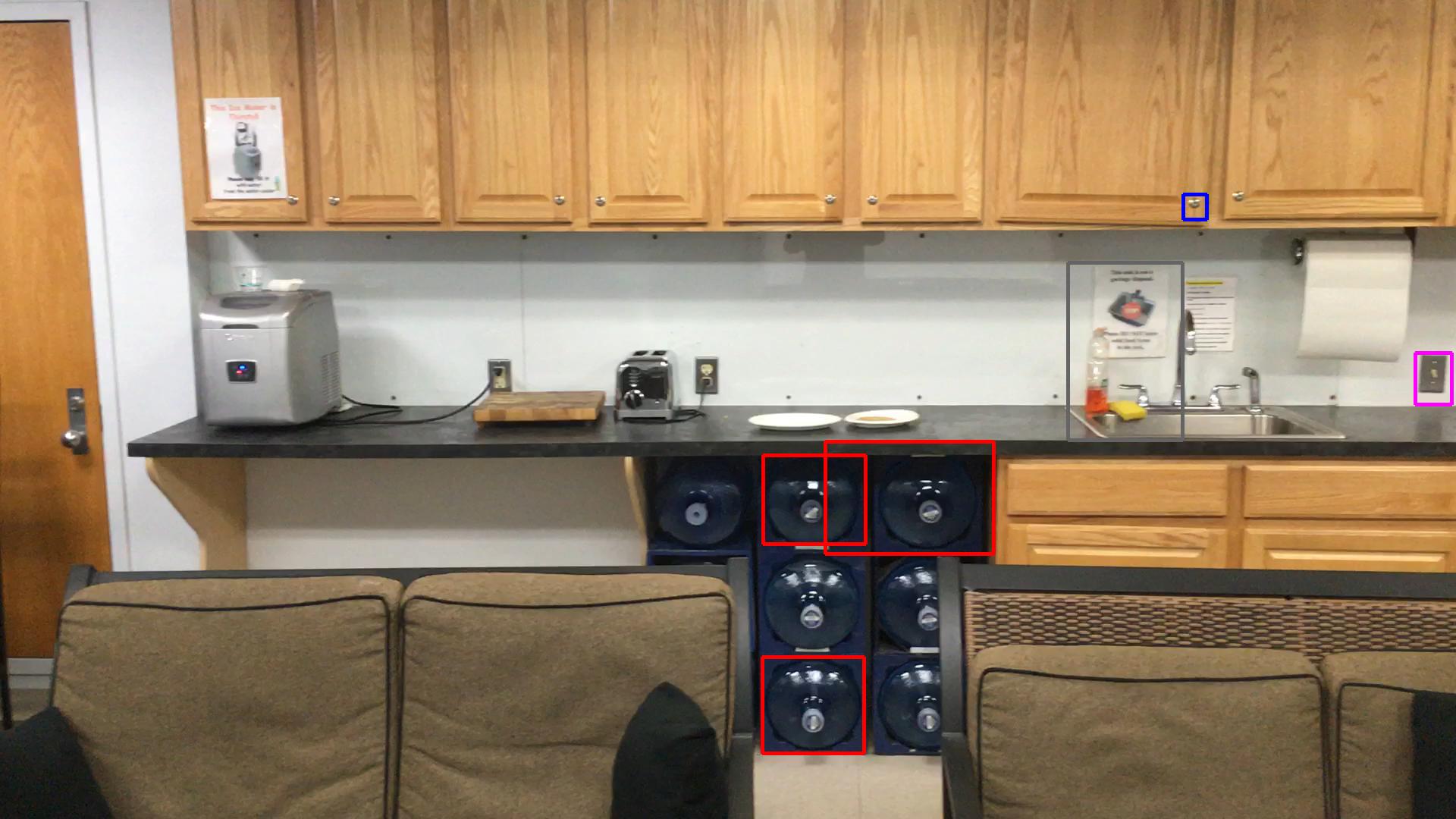}}
\subfigure[]{\includegraphics[width=2.1in]{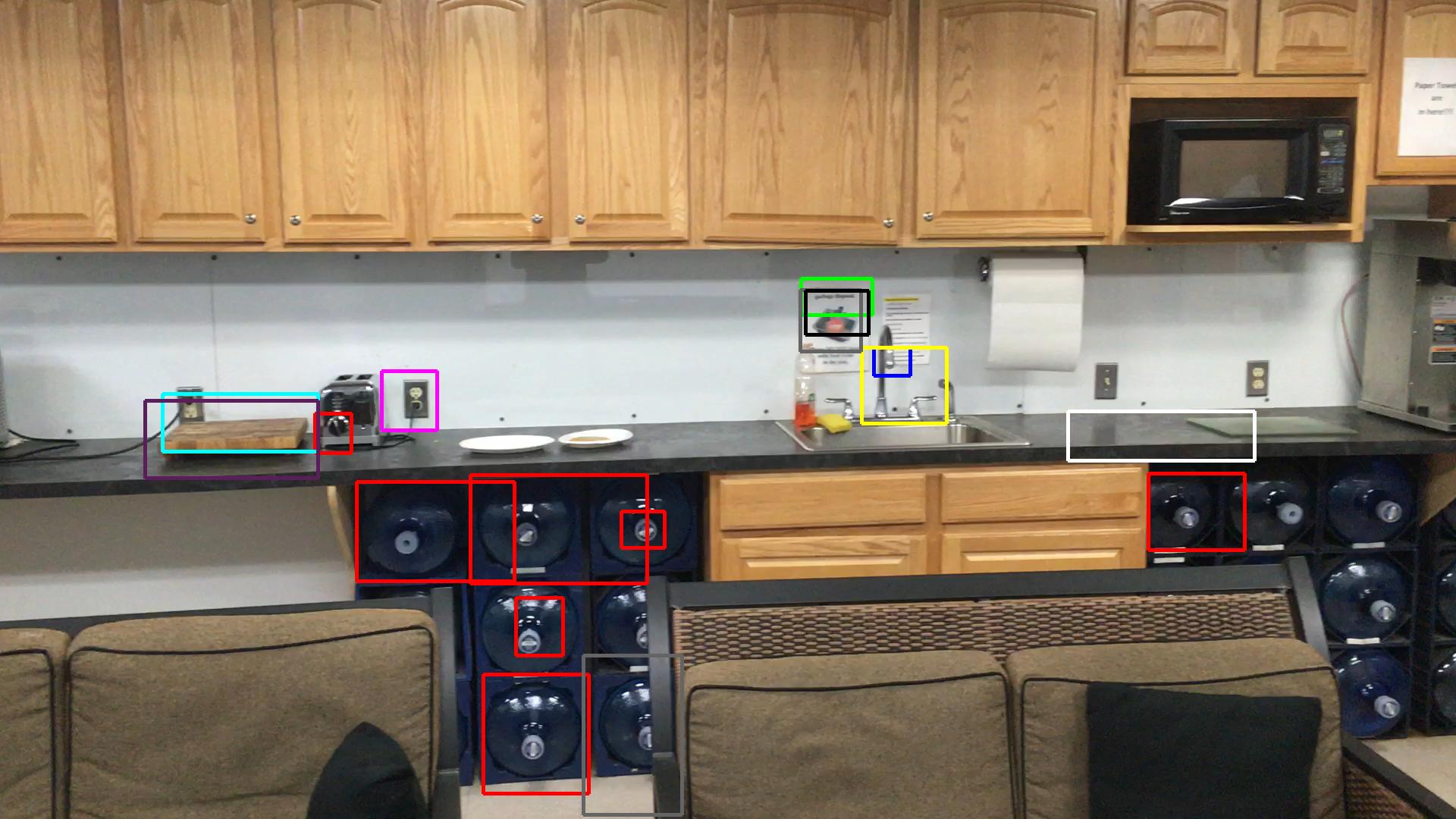}}
\subfigure[]{\includegraphics[width=2.1in]{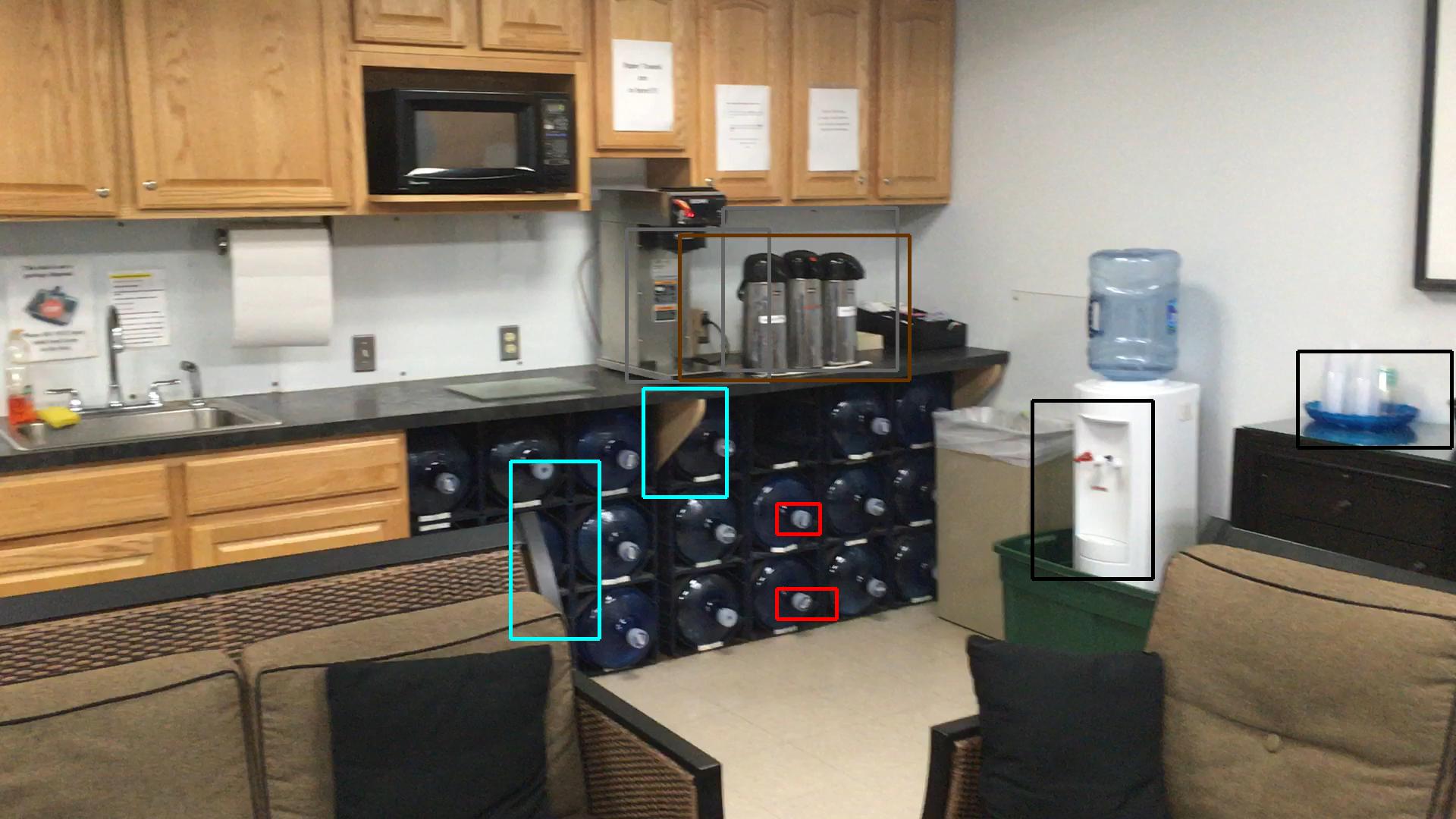}}
\subfigure[]{\includegraphics[width=2.1in]{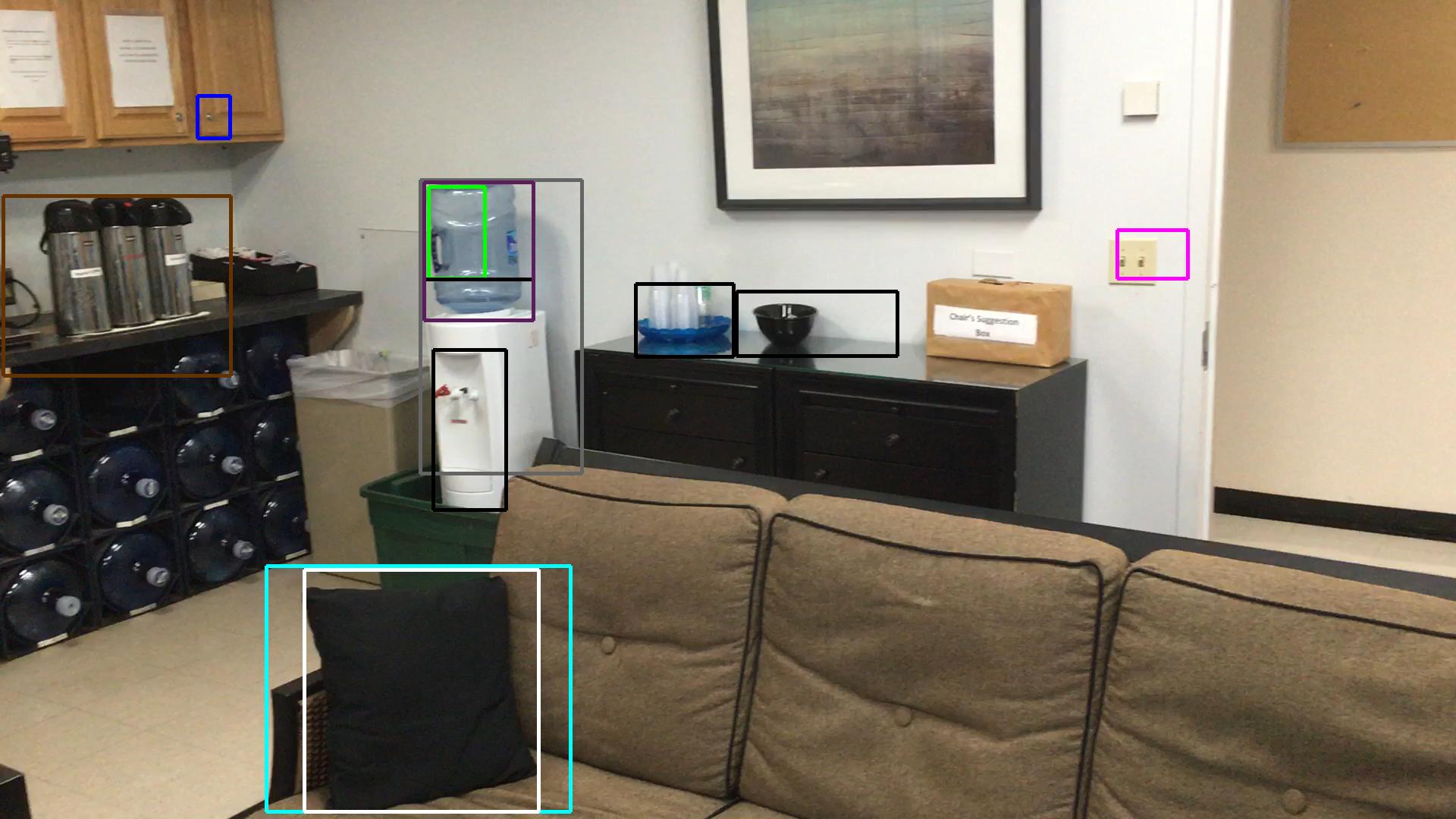}}
\caption{ Sample detection results of a dynamic kitchen scene using our system with the VGG-S network after three rounds of training.}
\label{VGG-S 3 video}
\end{figure*}

To further validate the applicability of the presented system, we deployed the trained model (VGG-S after three rounds of training) onto a humanoid platform (Baxter shown in Fig.~\ref{fg:kit} in a real world kitchen scene (a computer science student lounge). Fig.~\ref{VGG-S 3 video} lists example visualizations of our system's output. From the visualization,  we can see that the system successfully detects ``to sit or to place'' areas around chairs, ``turn on/off electricity'' around power outlets, ``turn on/off water'' areas around faucet and ``two hands raise'' areas around containers. We also notice an interesting failure case: the lids of the water bottles do not exist in the training data, but our system detects them as ``turn on fire''.  Even though the action consequence of turning the lid is predicted wrong, the intended movement (``turn to open'') is conceptually correct. These experimental results and observations also support our intuition that the trained functionality detection model has generalization capability and can be applied directly onto a humanoid platform. 

\section{Conclusion and Future Work}
\label{sec:conclusion}

In this work, we studied the problem of scene functional understanding for cognitive robots and presented a deep neural networks based two stage detection pipeline for the task. First of all, the problem itself is a brand new problem. We compiled a new scene functionality dataset from two existing publicly available dataset as testing beds and conducted experiments on it. The experimental results show our method is effective and efficient in detecting functional areas in an arbitrary indoor scene. We also conducted a cross validation experiment, and the experimental result supports our intuition that the presented functionality detection model has generalization potential. 

We found that due to the inherited ambiguity of functionality, the problem itself is more challenging than the traditional object recognition task. This partially explains why the experimental results still has potential space to improve. We also observed that during training stage, the classification accuracy is almost perfect on validation set, but during testing phase the detection performance is still far from perfect. One main reason for the performance drop is believed to be that the proposed regions contain novel variations or surroundings that the trained model fails to capture. It is generally considered as an open problem in vision tasks when using a general region proposal algorithm as the first stage of the detection pipeline. To tackle this problem, we are currently experimenting with one joint training approach, which aims to jointly train the region proposal and classification stages at the same time. 

Moreover, note that the functional categories we consider in the ontology are only direct functions, which are those can be directly applied onto objects, such as ``to open'' and ``to move''. It is much more complicated to categorize second-order or even third-order functions, such as ``to cut'', which requires first ``to move'' a tool (knife) and the function is then applied onto the second object (such as bread). This remains as an open problem for future research.


\newpage
\bibliographystyle{plainnat}
\bibliography{references}

\end{document}